\pdfoutput=1
\documentclass[11pt]{article}

\usepackage[final]{acl}

\usepackage{times}
\usepackage{latexsym}
\usepackage{enumitem}
\usepackage[T1]{fontenc}
\usepackage[utf8]{inputenc}
\usepackage{microtype}
\usepackage{inconsolata}
\usepackage{graphicx}
\usepackage{booktabs}
\usepackage{listings}
\usepackage{amsmath}
\usepackage{todonotes} 
\usepackage{titlesec}

\titlespacing*{\section}{0pt}{*0.7}{*0.7}
\titlespacing*{\subsection}{0pt}{*0.7}{*0.7}
\titlespacing*{\paragraph}{0pt}{*0.0}{*0.3}

\usepackage{url}
\title{Beyond IVR: Benchmarking Customer Support LLM Agents for Business-Adherence}

\author{
  Sumanth Balaji \quad Piyush Mishra \quad Aashraya Sachdeva\thanks{Corresponding author.} \quad Suraj Agrawal \\
  \texttt{\{sumanth.balaji, piyush.mishra, aashraya, suraj.agrawal\}@observe.ai} \\
  Observe.AI \\
  Bangalore, India
}

\begin{document}
\maketitle
\begin{abstract}
  Traditional customer support systems, such as Interactive Voice Response (IVR), rely on rigid scripts and lack the flexibility required for handling complex, policy-driven tasks. While large language model (LLM) agents offer a promising alternative, evaluating their ability to act in accordance with business rules and real-world support workflows remains an open challenge. Existing benchmarks primarily focus on tool usage or task completion, overlooking an agent's capacity to adhere to multi-step policies, navigate task dependencies, and remain robust to unpredictable user or environment behavior. In this work, we introduce JourneyBench, a benchmark designed to assess policy-aware agents in customer support. JourneyBench leverages graph representations to generate diverse, realistic support scenarios and proposes the User Journey Coverage Score, a novel metric to measure policy adherence. We evaluate multiple state-of-the-art LLMs using two agent designs: a Static-Prompt Agent (SPA) and a Dynamic-Prompt Agent (DPA) that explicitly models policy control. Across 703 conversations in three domains, we show that DPA significantly boosts policy adherence, even allowing smaller models like GPT-4o-mini to outperform more capable ones like GPT-4o. Our findings demonstrate the importance of structured orchestration and establish JourneyBench as a critical resource to advance AI-driven customer support beyond IVR-era limitations.

\end{abstract}
  
\section{Introduction}
\label{sec:introduction}
Customer support automation has traditionally relied on Interactive Voice Response (IVR) systems—automated telephone platforms that gather information and route calls through voice prompts and keypad inputs. While IVR enforces rigid flows via static decision trees to ensure compliance, it often lacks flexibility, resulting in poor user experience and high frustration \cite{Dean2008IVR, Coman2025IVR}. Advances in large language models (LLMs) enable LLM agents—autonomous systems combining textual reasoning and tool-use—to handle multi-turn conversations and dynamically manage customer support workflows \cite{react,schick2023toolformerlanguagemodelsteach,GuidedTOD}. Throughout this paper, we use ``agent'' to refer specifically to these LLM-powered autonomous systems. JourneyBench evaluates agents in text-based conversations, as extension to voice deployments is straightforward with speech-to-text and text-to-speech modules.


\setlength{\belowcaptionskip}{-15pt}
\begin{figure}[t]
    \centering
    \includegraphics[width=0.9\linewidth]{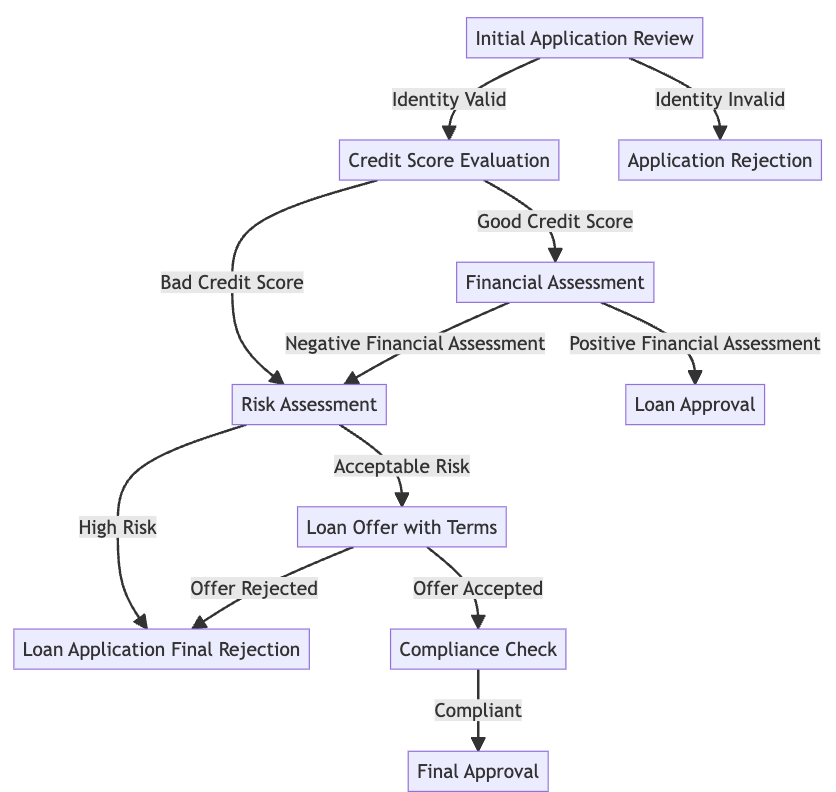}
    \caption{Example SOP graph for loan application processing, showing sequential tasks and decision points.}
    \label{fig:graph_example}
\end{figure}

Ensuring that agents follow business policies and procedural requirements remains a core challenge in production deployments. Standard Operating Procedures (SOPs) are structured workflows that prescribe execution order, validation checks, and exception handling protocols, encoding operational logic and compliance rules. As illustrated in Figure~\ref{fig:graph_example}, a compliant agent completes all required steps—identity verification, credit evaluation, financial assessment, risk assessment, and loan decision—with proper validations, whereas a non-compliant agent may skip risk assessment and proceed directly to approval, achieving the user’s goal while violating business logic and creating regulatory and financial risk. Existing benchmarks evaluate goal completion rather than pathway adherence, leaving this gap unaddressed. We therefore use the term \textbf{policy-aware agent} to denote an agent that consistently follows prescribed policies throughout the interaction.

\setlength{\belowcaptionskip}{-25pt}
\begin{figure}[t]
    \centering
    \includegraphics[width=0.6\linewidth]{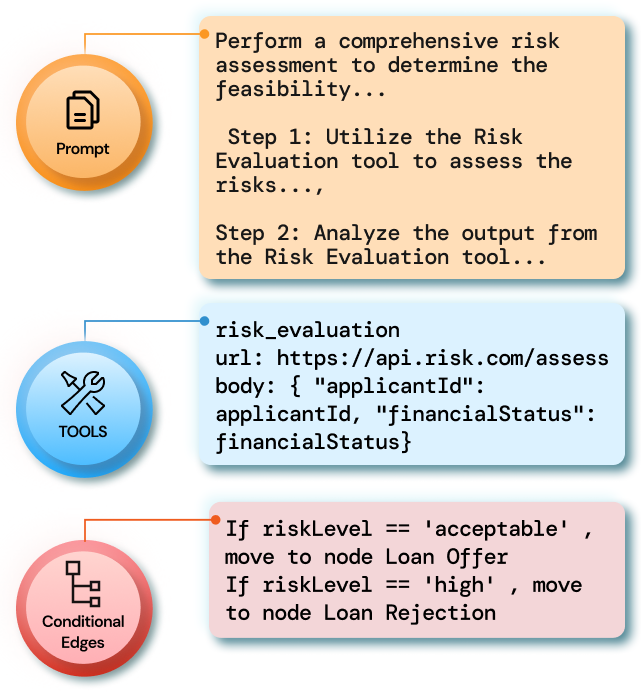}
    \caption{Components within a single node: the task description (prompt), available tools for execution, and conditional pathways (edges) that define transitions to the next node based on outcomes.}
    \label{fig:node_components}
\end{figure}

We distinguish \textbf{tools} (callable functions/APIs for atomic operations, e.g., \texttt{GET /customer/\{id\}}) from \textbf{tasks} (higher-level units combining multiple tools, e.g., ``identity verification''). Recent benchmarks \cite{tau-bench, toolsandbox, appworld} evaluate tool selection and state transitions, but inadequately assess complete task sequences with complex inter-task dependencies.

To address this gap, we introduce \textbf{JourneyBench}, a benchmark for evaluating \textbf{policy-aware agents} in customer support. JourneyBench represents SOPs as graphs to generate diverse scenarios, including challenges such as branching logic, missing inputs, and occasional tool failures. It also includes the User Journey Coverage Score (UJCS), which measures how well an agent follows the required sequence of actions defined by an SOP.

Our contributions are:
\begin{itemize}[nosep]
\item A benchmark, \textbf{JourneyBench}, for assessing policy-aware agents in customer support using graph-structured SOPs that capture task dependencies and policy constraints.
\item The \textbf{User Journey Coverage Score (UJCS)}, a metric for measuring adherence to SOP-mandated action sequences.
\item An empirical comparison showing that a \textbf{Dynamic-Prompt Agent} guided by workflow structure performs more reliably than a \textbf{Static-Prompt Agent}, highlighting the value of structured control in business settings.
\end{itemize}



\section{JourneyBench Framework}
\label{sec:journey_bench_framework}

\begin{figure*}[t]

    \centering
\includegraphics[width=0.8\linewidth, height=6.5cm, keepaspectratio]{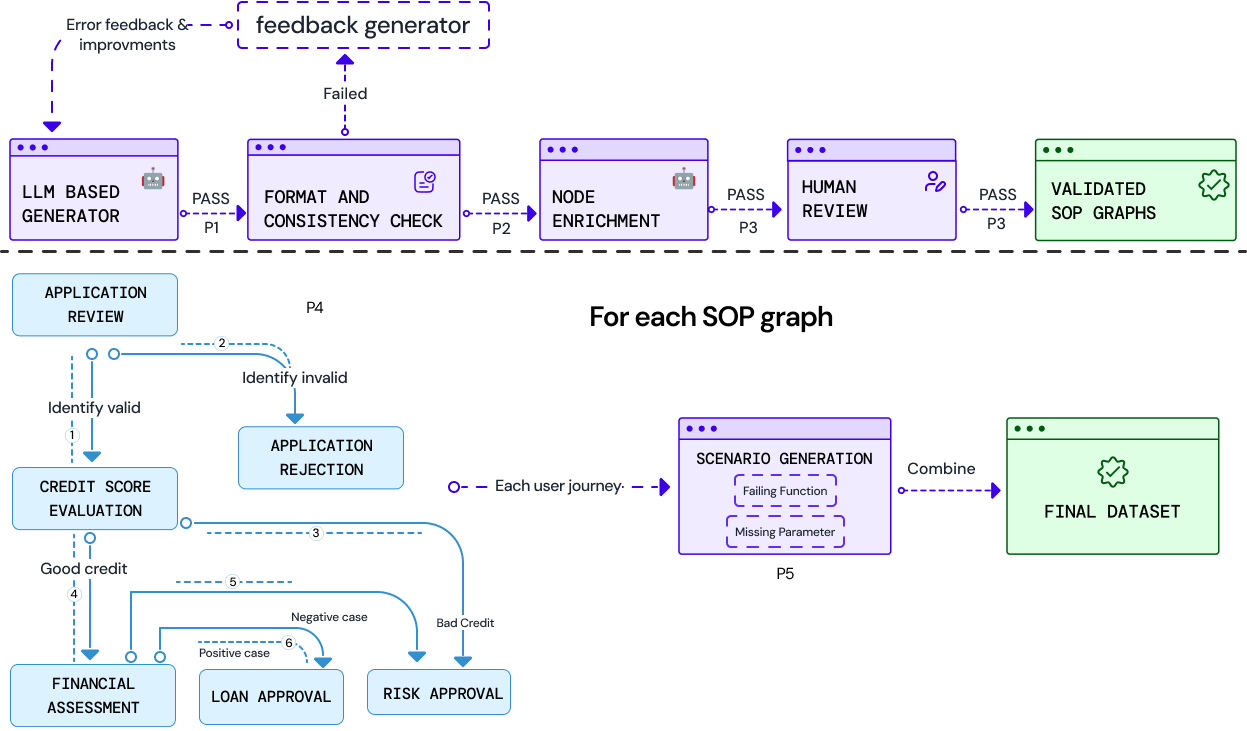}    \caption{
    Overview of our data generation pipeline across four phases (P1--P4). Validated SOP Graphs generate numbered user journeys via Breadth-First Search (BFS) traversal, which are then used to create diverse evaluation scenarios for the final benchmark dataset.
    }
    \label{fig:pipeline_overview}
\end{figure*}

The JourneyBench framework evaluates policy-aware agents using structured workflow representations. It consists of four core components: (1) SOP Graphs—Directed Acyclic Graphs encoding business workflows as tasks with conditional transitions; (2) Nodes—individual tasks with natural language descriptions, available tools, and procedural rules for state transitions; (3) User Journeys—specific paths through SOP graphs representing realistic agent-user interactions; and (4) Scenarios—test cases derived from user journeys that assess agent robustness under varied conditions, such as missing inputs or tool failures. Figure~\ref{fig:pipeline_overview} illustrates the multi-phase generation process.

\subsection{SOP Representation}
\label{sec:agent_graph_representation}
We model each SOP as a Directed Acyclic Graph (DAG), where nodes represent tasks and edges define valid transitions according to business logic. The DAG encodes task order, decision points, and policy constraints, serving as a blueprint for agent behavior. Figure~\ref{fig:graph_example} shows an example SOP graph for loan processing. Henceforth, we use ``node'' to refer to a task within the graph.


\begin{table}[t]
\centering
\resizebox{0.48\textwidth}{!}{
\begin{tabular}{lcccc}
\toprule
\textbf{Benchmark} & \textbf{Avg Turn} & \textbf{Avg Tool Calls} & \textbf{Dataset Size} & \textbf{Tools} \\
\midrule
\textbf{JourneyBench (ours)} & 10.91 & 3.34 & 703 & 41 \\
\quad \textit{E-commerce} & 13.37 & 3.06 & 232 & 12 \\
\quad \textit{Loan Application} & 6.57  & 3.69 & 230 & 15 \\
\quad \textit{Telecommunications} & 12.79 & 3.28 & 241 & 14 \\
\midrule
\textbf{TOOLSANDBOX}~\cite{toolsandbox} & 13.9  & 3.80 & 1032 & 34 \\
\textbf{BFCLV3}~\cite{BFCLV3}        & 2.00  & 0.78 & 2000 & 1193 \\
\textbf{Tau Bench}~\cite{tau-bench}  & 29.33 & 4.48 & 165  & 24 \\
\bottomrule
\end{tabular}
}
\caption{Comparison of JourneyBench with other agent benchmarks. JourneyBench statistics are presented overall and broken down by domain.}
\label{tab:benchmark_comparison}
\end{table}

\setlength{\belowcaptionskip}{-10pt}

\paragraph{Node Structure}
\label{sec:node_structure}
Each node represents a task, including its natural-language description, available tools (e.g., APIs), input/output parameters, and conditional pathways for transitions. Conditional pathways encode procedural rules as logical expressions over tool outputs (e.g., \texttt{riskLevel == 'acceptable'}), allowing complex workflow logic to be expressed clearly. At runtime, these conditions deterministically select the next node, ensuring strict adherence to the SOP. Figure~\ref{fig:node_components} illustrates a node's structure; technical details are in Appendix~\ref{app:node_example}.


\subsection{Synthetic Dataset Curation}
\label{sec:automated_graph_scenario_generation}
While SOP graphs can be manually authored for specific business processes, constructing a large and diverse benchmark requires a scalable generation pipeline. To build JourneyBench, we automatically generate a dataset of SOP graphs and corresponding interaction scenarios. To minimize human effort, we employ a multi-phase generation process inspired by recent work on LLM-based dataset creation\cite{barres2025tau2benchevaluatingconversationalagents}.

\paragraph{Phase 1: Graph Generation and Refinement} A state-of-the-art LLM generates foundational SOP graphs for 10 candidate domains, ensuring workflow complexity and realism. Outputs are validated for acyclicity and connectivity; if issues arise, an iterative LLM-based refinement resolves them (Appendix~\ref{app:graph_validation_details}). Once validated, node descriptions are enriched with detailed task/tool specifications and examples to reduce ambiguity. This method allows for creative, domain-specific workflows with reduced human effort (Appendix~\ref{app:graph_generation}).

\paragraph{Phase 2: Manual Review}
Human review ensures logical consistency in workflows, task and tool suitability, and overall graph quality. Each SOP graph is independently reviewed by five contact center agents (domain experts) against three binary pass/fail checks: \textbf{Logical Structure} (flow is logically correct and executable end-to-end), \textbf{Coherence} (node/tool descriptions and parameters are contextually appropriate and consistent), and \textbf{Complexity} (appropriate difficulty for the domain, neither trivial nor needlessly convoluted). A graph is accepted only if all five annotators unanimously pass all three checks ("5-of-5 agreement"). Of 10 candidate graphs, 4 met this standard; three diverse graphs—one per domain (Telecommunications, E-commerce, Loan Application)—were selected for benchmark experiments. This generate-and-filter approach enables rapid iteration: generating 10 diverse candidate graphs via LLM took under 1 hour, while manual authoring of comparable graphs would require weeks of expert time. Human review thus serves as a scalability multiplier rather than a bottleneck. See Appendix~\ref{app:human_validation} for details.

\paragraph{Phase 3: User Journey Generation}
\label{sec:benchmark_construction}
A user journey is a specific execution path through an SOP graph, representing the sequence of nodes and tool calls a user might follow to achieve their goal. We enumerate all possible paths using Breadth-First Search (BFS) (Figure~\ref{fig:pipeline_overview}).

Agents are evaluated via simulated conversations, with GPT-4o acting as the user, following established evaluation practices~\cite{tau-bench,toolsandbox}. Each simulation uses a \textbf{user seed}, a structured prompt specifying: (1) the target journey, (2) user information parameters (e.g., applicant ID), and (3) instructions for natural conversation through the tasks. Example seeds and templates are provided in Appendices~\ref{app:user_seed_example} and~\ref{app:user_simulation_template}.



JourneyBench evaluates workflow adherence rather than tool implementation, treating tools as black boxes with pre-generated responses. For each journey, tool responses that influence workflow branching (e.g., \texttt{riskLevel}) are set algorithmically to follow the target path (Appendix~\ref{app:value_generation_algorithm}), while other outputs, such as timestamps or confirmation IDs, are generated by an LLM for realism. During evaluation, agents receive these pre-generated responses, ensuring deterministic and reproducible testing. All user journeys are manually reviewed for logical consistency before scenario generation.



\paragraph{Phase 4: Scenario Data Generation}
From each user journey, we generate multiple evaluation \textbf{scenarios}, each representing a full conversational test case with an initial state and expected outcome. The baseline is the “correct context” scenario, where all user parameters are present and tools work as intended. From each correct context case, we systematically construct two additional scenario types: \textbf{Missing Parameter}, where required user inputs are withheld and unreachable tool calls are removed from the expected tool trace; and \textbf{Failing Function}, where a tool call fails (e.g., API error), and the trace is updated to remove downstream calls that can no longer execute. Duplicate scenarios with identical sequences and responses are removed to ensure uniqueness.

\setlength{\belowcaptionskip}{-15pt}
\begin{table*}[t]
\centering
\resizebox{\textwidth}{!}{%
\begin{tabular}{lccc|ccc}
\toprule
& \multicolumn{3}{c|}{SPA\textsuperscript{*}} 
    & \multicolumn{3}{c}{DPA\textsuperscript{†}} \\
\midrule
\textbf{Model}
& \textbf{Correct Context} & \textbf{Failing Function} & \textbf{Missing Parameter} 
    & \textbf{Correct Context} & \textbf{Failing Function} & \textbf{Missing Parameter} \\
    \midrule
GPT-4o~\cite{gpt4o}           & 0.871 & 0.511 & 0.309 & 0.873 & 0.857 & 0.530 \\
GPT-4o-mini~\cite{gpt4o}      & 0.720 & 0.326 & 0.263 & 0.718 & 0.816 & 0.414 \\
Claude 3.5 Haiku~\cite{anthropic2024claude35haiku} & 0.234 & 0.285 & 0.240 & 0.504 & 0.776 & 0.453 \\
Llama 3.3~\cite{llama3}        & 0.237 & 0.264 & 0.256 & 0.311 & 0.345 & 0.332 \\
\bottomrule
\end{tabular}%
}
\caption{User Journey Coverage Scores (UJCS) for Dynamic-Prompt-Agent (DPA\textsuperscript{†}) and Static-Prompt-Agent (SPA\textsuperscript{*}) across scenario types. Higher the better.}
\label{tab:overall_performance}
\end{table*}

Compared to other benchmarks, JourneyBench offers significant conversational depth (10.91 turns, 3.34 tool calls per conversation), a unique toolset (41 tools), and a large dataset (703 datapoints), providing a robust benchmark for testing agentic capabilities in policy-driven domains (See Table~\ref{tab:benchmark_comparison}).

\section{Evaluation Metrics}
\label{sec:evaluation_metrics}
To assess an agent's adherence to business workflows, we evaluate its performance on simulated conversational scenarios from user journeys. Each journey specifies a sequence of tool calls and parameters, so our evaluation checks strict procedural adherence and execution accuracy. This forms the basis of our main metric, the User Journey Coverage Score (UJCS).

\paragraph{Tool Trace Alignment}: 
For each simulated conversation, Tool Trace Alignment compares the predicted tool call sequence ($T_{act}$) with the expected sequence ($T_{exp}$). Any missing, extra, or misordered tool call indicates an SOP violation, giving the conversation a score of 0.

\paragraph{Tool Call Accuracy}: For each simulated conversation, this metric quantifies the correctness of parameter values supplied during tool execution. Tool Call Accuracy $TCA_{conv}$, score for a single conversation is defined by Equation~\ref{eq:s_conv} where $C_i = |P_{act}^{(i)} \cap P_{exp}^{(i)}|$ is the count of correct parameters for the $i$-th tool call, and $E_i = |P_{exp}^{(i)}|$ is the number of expected parameters, $L = |T_{exp}|$ is the length of the trace, and $S_{conv}$, score for a single conversation.

\setlength{\belowcaptionskip}{-15pt}
\begin{equation} \label{eq:s_conv}
TCA_{conv} =
\begin{cases}
\frac{\sum_{i=1}^{L} C_i}{\sum_{i=1}^{L} E_i} & \text{if } T_{act} = T_{exp} \\
0 & \text{otherwise}
\end{cases}
\end{equation}

\paragraph{User Journey Coverage Score (UJCS)}: The metric evaluates overall efficacy of an agent for a given SOP graph on N conversations is. 

\begin{equation}\label{eq:journey_coverage}
    \text{UJCS} = \frac{1}{N} \sum_{k=1}^{N} TCA_{conv_k}
\end{equation}

\section{Experimentation}
\label{sec:agent_workflow_orchestration_approaches}

\subsection{Instantiating Agents with SOP}
To study agent adherence to SOPs, we instantiate two variants of agents: 

\paragraph{Static-Prompt-Agent (SPA):} SPA uses a single, static system prompt. Using a consistent textual template, the entire SOP is transformed into one comprehensive system prompt. The SOP's conditional branching logic is encoded using if-then statements. Tools for all nodes are added to the system prompt as well. An example of this prompt structure can be found in Appendix~\ref{app:prompt_agent_example_prompt}.

\paragraph{Dynamic-Prompt-Agent (DPA):} The DPA models the SOP as a state machine, processing one node at a time (see Appendix~\ref{app:node_example}). After each tool execution, an orchestrator—a control component that manages the workflow state and transitions, interprets the response (Appendix~\ref{app:tool_response_example}), and determines the next node by evaluating conditional pathways defined in the SOP logic. Each transition replaces the previous prompt and updates the accessible tools. This design minimizes context overload, supports mid-flow corrections, and promotes reliable policy execution (see Appendix~\ref{app:ivr_vs_dpa}).

We exclude explicit planning-based approaches such as ReAct\cite{react} due to their significant latency, which makes them unsuitable for real-time interactions in customer support. Additionally, we developed a custom framework for the management of SOP's runtime state and facilitate the handling of conditional pathways. Popular libraries (e.g., LangGraph, CrewAI \cite{LangGraph2023,CrewAI2023}) could help construct agents of similar capabilities; we chose a custom framework to ensure reproducibility and stable experimental control across runs and to avoid dependency churn.

\subsection{Experiments}
\label{sec:experiments}
All experiments use a 40-turn limit and default LLM temperature settings. The simulated user (GPT-4o, Section~\ref{sec:benchmark_construction}) follows the predefined journey while maintaining natural conversation and preventing information leakage. From each correct-context journey, our benchmark generates additional scenarios with failed functions or missing parameters to test agent robustness.


\begin{table*}[t]
\centering
\resizebox{\textwidth}{!}{%
\begin{tabular}{lccc|c|ccc|c}
\toprule
& 
  \multicolumn{4}{c|}{SPA\textsuperscript{*}} 
    & \multicolumn{4}{c}{DPA\textsuperscript{†}} \\
\midrule
\textbf{Model} 
 & \textbf{E-commerce} & \textbf{Loan Application} & \textbf{Telecommunications} 
    & \textbf{Average} 
    & \textbf{E-commerce} & \textbf{Loan Application} & \textbf{Telecommunications} 
    & \textbf{Average} \\
\midrule
GPT-4o~\cite{gpt4o}           & 0.617 & 0.651 & 0.423 & \textbf{0.564} 
                               & 0.730 & 0.776 & 0.646 & \textbf{0.717} \\
GPT-4o-mini~\cite{gpt4o}      & 0.502 & 0.504 & 0.304 & 0.437 
                               & 0.679 & 0.623 & 0.646 & \textbf{0.649} \\
Claude 3.5 Haiku~\cite{anthropic2024claude35haiku} & 0.359 & 0.286 & 0.116 & 0.253 
                                                  & 0.593 & 0.615 & 0.525 & 0.578 \\
Llama 3.3~\cite{llama3}        & 0.360 & 0.278 & 0.119 & 0.252 
                               & 0.432 & 0.329 & 0.228 & 0.330 \\
\bottomrule
\end{tabular}%
}
\caption{User Journey Coverage Scores (UJCS) for Dynamic-Prompt-Agent (DPA\textsuperscript{†}) and Static-Prompt-Agent (SPA\textsuperscript{*}) across customer service domains. Higher the better.}
\label{tab:domain_breakdown}
\end{table*}

\paragraph{Metrics:}
Agent performance is evaluated using the \textbf{User Journey Coverage Score} (refer to  Section~\ref{sec:evaluation_metrics}). We also track the number of successfully completed conversations and various error types.

\paragraph{Real-World Deployment:} The structured DPA-based orchestration is deployed in production across client contact centers, reliably handling 6{,}000+ calls daily while meeting real-time and policy-adherence requirements. These production systems process voice calls by converting speech to text, applying the same text-based DPA workflow logic evaluated in JourneyBench, and converting responses back to speech. This operational footprint demonstrates that structured agent control is practical and effective beyond controlled simulations.

\paragraph{Realism Validation (LLM-as-a-Judge):} 
To ensure synthetic conversations reflect production-quality interactions, we evaluate them using the same LLM-as-a-judge rubric applied in client Quality Assurance (QA). The rubric measures \textit{Conversational Proficiency} (CP; e.g., empathy, clarity, turn-taking) and \textit{Goal Attainment} (GA; e.g., intent recognition, request resolution) via binary Yes/No questions aggregated across conversations. Synthetic conversations achieve 84.37\% overall (82.33\% CP; 87.78\% GA), comparable to production QA distributions, indicating that benchmark traces realistically capture agent behavior and policy adherence. Appendix~\ref{app:realism_validation} details rubric based validation.


\paragraph{Results and Analysis:}
\label{sec:results}
Evaluations on \textbf{JourneyBench} demonstrate consistent performance gains with the \textbf{Dynamic-Prompt-Agent (DPA)} over the \textbf{Static-Prompt-Agent (SPA)}. As shown in Tables~\ref{tab:overall_performance} and \ref{tab:domain_breakdown}, GPT-4o with DPA achieves a UJCS of \textbf{0.717}, substantially higher than SPA’s 0.564, highlighting the value of explicit workflow guidance for policy adherence. Scenario-based testing further shows that SPA performance drops under disturbances such as failed functions or missing parameters, whereas DPA maintains stable coverage across all scenarios. Notably, GPT-4o-mini with DPA (0.649) outperforms GPT-4o with SPA (0.564), demonstrating that structured orchestration enables smaller, cost-efficient models to match or exceed larger ones.



\subsection{Error Analysis}
\label{sec:error_analysis}
We manually went through conversations where UJCS was low to identify error classes. We group the errors into the following three classes:

\paragraph{Dependency Violations:}
Dependency violations occur when an agent proceeds without required parameters or prior tool use, violating SOP logic. \textbf{SPA} often advanced despite missing inputs or failures, while \textbf{DPA} correctly halted to maintain logical consistency. More examples are in Appendix~\ref{app:dependency_violations}.

\paragraph{Hallucination in Parameter Values:}
Parameter hallucination occurs when an agent uses example values from a tool description instead of the user’s input, leading to incorrect tool usage. For example, a user credit score of 720 might be replaced by 700 from the tool description. See Appendix~\ref{app:hallucinations} for an example. Both \textbf{SPA} and \textbf{DPA} showed this behavior, though DPA was less prone due to node-specific tool restrictions.

\paragraph{User Simulator Failures:}
We observed failures from the LLM-based user simulator, which do not reflect agent performance but can affect evaluation reliability. JourneyBench helped identify two issues: \textbf{user input hallucination}, where the simulator provides info not in the user seed, and \textbf{incomplete user journeys}, where the conversation ends prematurely before the required journey is completed(Appendix~\ref{app:collusion_example},~\ref{app:user_mistake}).

\section{Related Work}
\label{sec:related}
Recent literature has explored evaluating LLMs in multi-turn, tool-use settings. Benchmarks like Tau Bench~\cite{tau-bench} pioneered simulation-driven evaluation, and its successor Tau2-Bench~\cite{barres2025tau2benchevaluatingconversationalagents} extended this to dual-control environments. Others, like ToolSandBox~\cite{toolsandbox} and AppWorld~\cite{appworld}, focus on stateful execution and world-state tracking. While these frameworks advanced agent evaluation, they primarily measure tool selection accuracy or state changes, not an agent's fidelity to a prescribed, multi-step journey with complex dependencies, a gap JourneyBench addresses. For instance, BFCLV3~\cite{BFCLV3} uses static conversations but does not test agent responses to dynamic tool failures. Our work, in contrast, specifically evaluates policy adherence, task dependencies, and robustness to common conversational disturbances.

Furthermore, a common theme in existing benchmarks is the need to manually define tool logic and database states. This approach poses a challenge to scale to production environments where tools and data constantly evolve. Following principles of separation of concerns from system design, JourneyBench treats tools as modular components with well-defined interfaces, decoupling their internal implementation from workflow evaluation. This design allows our benchmark to focus on an agent's adherence to workflow logic rather than tool-specific behavior, a key distinction from prior work.
\section{Conclusion}
\label{sec:conclusion}
Moving customer support beyond rigid IVR systems requires agents that combine conversational flexibility with strict policy adherence—a capability existing benchmarks fail to measure. We introduced JourneyBench, a benchmark that evaluates policy-aware agents through graph-based SOP representations and the User Journey Coverage Score metric. Across 703 conversations spanning three domains, we demonstrated that structured workflow orchestration (Dynamic-Prompt-Agent) significantly outperforms prompt-based approaches (Static-Prompt-Agent), enabling even smaller models to exceed larger ones in policy compliance. Our approach is validated in production, where DPA-based systems reliably handle 6,000+ daily customer interactions. By providing both rigorous evaluation methodology and evidence that structured control enables robust, policy-compliant automation, JourneyBench establishes a foundation for deploying reliable AI agents in high-stakes business environments.
\section{Limitations and Future Work}
Our framework shows strong utility but has limitations that suggest avenues for future research. The \textbf{Dynamic-Prompt-Agent}'s success depends on precise modeling of business logic, which can be challenging in dynamic or poorly documented fields. Future work might explore semi-automated graph generation from conversation logs. Our simulation-based evaluation may not capture all nuances of real-world user behavior, and the high cost (\$388.88) constrained the range of models we tested. Future research could focus on more cost-effective evaluation methods and complex dependency structures.

\section{Ethical Considerations}
\label{sec:ethical_considerations}
The use of synthetically generated benchmarks raises important considerations that we address through our methodology and recommend for practitioners.

\paragraph{Synthetic Data Quality:}
As JourneyBench uses LLMs to generate workflows and conversations, it may inherit model biases. We mitigate this through domain-expert validation (Section~\ref{sec:automated_graph_scenario_generation}) and QA-based checks of conversational realism (Section~\ref{sec:experiments}). Organization benchmarks should be paired with real-world bias checks in deployed systems.


\paragraph{Evaluation Validity:}
LLM-generated evaluation data can introduce circularity. JourneyBench limits this risk by evaluating adherence to human-defined SOP structures rather than free-form generation. Human validation and alignment with production behavior provide additional grounding. We recommend using JourneyBench alongside human assessment.


\paragraph{Workforce Impact:}
Customer support automation can affect staffing. Our deployments suggest a shift toward higher-complexity tasks rather than direct displacement, but organizations should plan responsible transitions and training.


\bibliography{custom}

@article{Dean2008IVR,
  author = "Dean, D.H.",
  title = "What's wrong with IVR self‐service",
  journal = "Managing Service Quality: An International Journal",
  volume = "18",
  number = "6",
  pages = "594--609",
  year = "2008",
  doi = "10.1108/09604520810920086",
  url = "https://doi.org/10.1108/09604520810920086"
}

@misc{schick2023toolformerlanguagemodelsteach,
      title={Toolformer: Language Models Can Teach Themselves to Use Tools}, 
      author={Timo Schick and Jane Dwivedi-Yu and Roberto Dessì and Roberta Raileanu and Maria Lomeli and Luke Zettlemoyer and Nicola Cancedda and Thomas Scialom},
      year={2023},
      eprint={2302.04761},
      archivePrefix={arXiv},
      primaryClass={cs.CL},
      url={https://arxiv.org/abs/2302.04761}, 
}

@article{Coman2025IVR,
  title = "IVR systems used in call center management: a scientometric analysis of the literature",
  author = "Coman, Ecaterina",
  journal = "Frontiers in Computer Science",
  volume = "7",
  year = "2025",
  issn = "2624-9898",
  url = "https://www.frontiersin.org/journals/computer-science/articles/10.3389/fcomp.2025.1459787",
  doi = "10.3389/fcomp.2025.1459787"
}

@inproceedings{GuidedTOD,
    title = "Guideline Compliance in Task-Oriented Dialogue: The Chained Prior Approach",
    author = "Wen, Xiangyu  and
      Zhong, Jianyuan  and
      Xu, Zhijian  and
      Xu, Qiang",
    editor = "Chiruzzo, Luis  and
      Ritter, Alan  and
      Wang, Lu",
    booktitle = "Findings of the Association for Computational Linguistics: NAACL 2025",
    month = apr,
    year = "2025",
    address = "Albuquerque, New Mexico",
    publisher = "Association for Computational Linguistics",
    url = "https://aclanthology.org/2025.findings-naacl.377/",
    pages = "6750--6776",
    ISBN = "979-8-89176-195-7"
}

@misc{react,
      title={ReAct: Synergizing Reasoning and Acting in Language Models}, 
      author={Shunyu Yao and Jeffrey Zhao and Dian Yu and Nan Du and Izhak Shafran and Karthik Narasimhan and Yuan Cao},
      year={2023},
      eprint={2210.03629},
      archivePrefix={arXiv},
      primaryClass={cs.CL},
      url={https://arxiv.org/abs/2210.03629}, 
}

@misc{tau-bench,
      title={$\tau$-bench: A Benchmark for Tool-Agent-User Interaction in Real-World Domains}, 
      author={Shunyu Yao and Noah Shinn and Pedram Razavi and Karthik Narasimhan},
      year={2024},
      eprint={2406.12045},
      archivePrefix={arXiv},
      primaryClass={cs.AI},
      url={https://arxiv.org/abs/2406.12045}, 
}

@misc{LangGraph2023,
  author = "{LangChain AI}",
  title = "LangGraph",
  year = "2023",
  howpublished = "\url{https://github.com/langchain-ai/langgraph}"
}

@misc{barres2025tau2benchevaluatingconversationalagents,
      title={$\tau^2$-Bench: Evaluating Conversational Agents in a Dual-Control Environment}, 
      author={Victor Barres and Honghua Dong and Soham Ray and Xujie Si and Karthik Narasimhan},
      year={2025},
      eprint={2506.07982},
      archivePrefix={arXiv},
      primaryClass={cs.AI},
      url={https://arxiv.org/abs/2506.07982}, 
}

@misc{CrewAI2023,
  author = "{CrewAI Inc.}",
  title = "CrewAI",
  year = "2023",
  howpublished = "\url{https://github.com/crewAIInc/crewAI}",
}

@inproceedings{BFCLV3,

        title={Berkeley Function Calling Leaderboard},

        author={Fanjia Yan and Huanzhi Mao and Charlie Cheng-Jie Ji and Tianjun Zhang and Shishir G. Patil and Ion Stoica and Joseph E. Gonzalez},

        year={2024},

        howpublished={\url{https://gorilla.cs.berkeley.edu/blogs/8_berkeley_function_calling_leaderboard.html}},

    }

@misc{AppWorld,
      title={AppWorld: A Controllable World of Apps and People for Benchmarking Interactive Coding Agents}, 
      author={Harsh Trivedi and Tushar Khot and Mareike Hartmann and Ruskin Manku and Vinty Dong and Edward Li and Shashank Gupta and Ashish Sabharwal and Niranjan Balasubramanian},
      year={2024},
      eprint={2407.18901},
      archivePrefix={arXiv},
      primaryClass={cs.SE},
      url={https://arxiv.org/abs/2407.18901}, 
}

@misc{ToolSandBox,
      title={ToolSandbox: A Stateful, Conversational, Interactive Evaluation Benchmark for LLM Tool Use Capabilities}, 
      author={Jiarui Lu and Thomas Holleis and Yizhe Zhang and Bernhard Aumayer and Feng Nan and Felix Bai and Shuang Ma and Shen Ma and Mengyu Li and Guoli Yin and Zirui Wang and Ruoming Pang},
      year={2025},
      eprint={2408.04682},
      archivePrefix={arXiv},
      primaryClass={cs.CL},
      url={https://arxiv.org/abs/2408.04682}, 
}

@misc{gpt4o,
      title={GPT-4o: Visual perception performance of multimodal large language models in piglet activity understanding}, 
      author={Yiqi Wu and Xiaodan Hu and Ziming Fu and Siling Zhou and Jiangong Li},
      year={2024},
      eprint={2406.09781},
      archivePrefix={arXiv},
      primaryClass={cs.CV},
      url={https://arxiv.org/abs/2406.09781}, 
}

@misc{anthropic2024claude35haiku,
  title     = {Claude 3.5 Haiku},
  author    = {Anthropic},
  year      = {2024},
  howpublished = {\url{https://www.anthropic.com/claude/haiku}},
  note      = {Accessed: 2025-05-18}
}

@misc{llama3,
      title={The Llama 3 Herd of Models}, 
      author={Aaron Grattafiori and Abhimanyu Dubey and Abhinav Jauhri and Abhinav Pandey and Abhishek Kadian and Ahmad Al-Dahle and Aiesha Letman and Akhil Mathur and Alan Schelten and Alex Vaughan and Amy Yang and Angela Fan and Anirudh Goyal and Anthony Hartshorn and Aobo Yang and Archi Mitra and Archie Sravankumar and Artem Korenev and Arthur Hinsvark and Arun Rao and Aston Zhang and Aurelien Rodriguez and Austen Gregerson and Ava Spataru and Baptiste Roziere and Bethany Biron and Binh Tang and Bobbie Chern and Charlotte Caucheteux and Chaya Nayak and Chloe Bi and Chris Marra and Chris McConnell and Christian Keller and Christophe Touret and Chunyang Wu and Corinne Wong and Cristian Canton Ferrer and Cyrus Nikolaidis and Damien Allonsius and Daniel Song and Danielle Pintz and Danny Livshits and Danny Wyatt and David Esiobu and Dhruv Choudhary and Dhruv Mahajan and Diego Garcia-Olano and Diego Perino and Dieuwke Hupkes and Egor Lakomkin and Ehab AlBadawy and Elina Lobanova and Emily Dinan and Eric Michael Smith and Filip Radenovic and Francisco Guzmán and Frank Zhang and Gabriel Synnaeve and Gabrielle Lee and Georgia Lewis Anderson and Govind Thattai and Graeme Nail and Gregoire Mialon and Guan Pang and Guillem Cucurell and Hailey Nguyen and Hannah Korevaar and Hu Xu and Hugo Touvron and Iliyan Zarov and Imanol Arrieta Ibarra and Isabel Kloumann and Ishan Misra and Ivan Evtimov and Jack Zhang and Jade Copet and Jaewon Lee and Jan Geffert and Jana Vranes and Jason Park and Jay Mahadeokar and Jeet Shah and Jelmer van der Linde and Jennifer Billock and Jenny Hong and Jenya Lee and Jeremy Fu and Jianfeng Chi and Jianyu Huang and Jiawen Liu and Jie Wang and Jiecao Yu and Joanna Bitton and Joe Spisak and Jongsoo Park and Joseph Rocca and Joshua Johnstun and Joshua Saxe and Junteng Jia and Kalyan Vasuden Alwala and Karthik Prasad and Kartikeya Upasani and Kate Plawiak and Ke Li and Kenneth Heafield and Kevin Stone and Khalid El-Arini and Krithika Iyer and Kshitiz Malik and Kuenley Chiu and Kunal Bhalla and Kushal Lakhotia and Lauren Rantala-Yeary and Laurens van der Maaten and Lawrence Chen and Liang Tan and Liz Jenkins and Louis Martin and Lovish Madaan and Lubo Malo and Lukas Blecher and Lukas Landzaat and Luke de Oliveira and Madeline Muzzi and Mahesh Pasupuleti and Mannat Singh and Manohar Paluri and Marcin Kardas and Maria Tsimpoukelli and Mathew Oldham and Mathieu Rita and Maya Pavlova and Melanie Kambadur and Mike Lewis and Min Si and Mitesh Kumar Singh and Mona Hassan and Naman Goyal and Narjes Torabi and Nikolay Bashlykov and Nikolay Bogoychev and Niladri Chatterji and Ning Zhang and Olivier Duchenne and Onur Çelebi and Patrick Alrassy and Pengchuan Zhang and Pengwei Li and Petar Vasic and Peter Weng and Prajjwal Bhargava and Pratik Dubal and Praveen Krishnan and Punit Singh Koura and Puxin Xu and Qing He and Qingxiao Dong and Ragavan Srinivasan and Raj Ganapathy and Ramon Calderer and Ricardo Silveira Cabral and Robert Stojnic and Roberta Raileanu and Rohan Maheswari and Rohit Girdhar and Rohit Patel and Romain Sauvestre and Ronnie Polidoro and Roshan Sumbaly and Ross Taylor and Ruan Silva and Rui Hou and Rui Wang and Saghar Hosseini and Sahana Chennabasappa and Sanjay Singh and Sean Bell and Seohyun Sonia Kim and Sergey Edunov and Shaoliang Nie and Sharan Narang and Sharath Raparthy and Sheng Shen and Shengye Wan and Shruti Bhosale and Shun Zhang and Simon Vandenhende and Soumya Batra and Spencer Whitman and Sten Sootla and Stephane Collot and Suchin Gururangan and Sydney Borodinsky and Tamar Herman and Tara Fowler and Tarek Sheasha and Thomas Georgiou and Thomas Scialom and Tobias Speckbacher and Todor Mihaylov and Tong Xiao and Ujjwal Karn and Vedanuj Goswami and Vibhor Gupta and Vignesh Ramanathan and Viktor Kerkez and Vincent Gonguet and Virginie Do and Vish Vogeti and Vítor Albiero and Vladan Petrovic and Weiwei Chu and Wenhan Xiong and Wenyin Fu and Whitney Meers and Xavier Martinet and Xiaodong Wang and Xiaofang Wang and Xiaoqing Ellen Tan and Xide Xia and Xinfeng Xie and Xuchao Jia and Xuewei Wang and Yaelle Goldschlag and Yashesh Gaur and Yasmine Babaei and Yi Wen and Yiwen Song and Yuchen Zhang and Yue Li and Yuning Mao and Zacharie Delpierre Coudert and Zheng Yan and Zhengxing Chen and Zoe Papakipos and Aaditya Singh and Aayushi Srivastava and Abha Jain and Adam Kelsey and Adam Shajnfeld and Adithya Gangidi and Adolfo Victoria and Ahuva Goldstand and Ajay Menon and Ajay Sharma and Alex Boesenberg and Alexei Baevski and Allie Feinstein and Amanda Kallet and Amit Sangani and Amos Teo and Anam Yunus and Andrei Lupu and Andres Alvarado and Andrew Caples and Andrew Gu and Andrew Ho and Andrew Poulton and Andrew Ryan and Ankit Ramchandani and Annie Dong and Annie Franco and Anuj Goyal and Aparajita Saraf and Arkabandhu Chowdhury and Ashley Gabriel and Ashwin Bharambe and Assaf Eisenman and Azadeh Yazdan and Beau James and Ben Maurer and Benjamin Leonhardi and Bernie Huang and Beth Loyd and Beto De Paola and Bhargavi Paranjape and Bing Liu and Bo Wu and Boyu Ni and Braden Hancock and Bram Wasti and Brandon Spence and Brani Stojkovic and Brian Gamido and Britt Montalvo and Carl Parker and Carly Burton and Catalina Mejia and Ce Liu and Changhan Wang and Changkyu Kim and Chao Zhou and Chester Hu and Ching-Hsiang Chu and Chris Cai and Chris Tindal and Christoph Feichtenhofer and Cynthia Gao and Damon Civin and Dana Beaty and Daniel Kreymer and Daniel Li and David Adkins and David Xu and Davide Testuggine and Delia David and Devi Parikh and Diana Liskovich and Didem Foss and Dingkang Wang and Duc Le and Dustin Holland and Edward Dowling and Eissa Jamil and Elaine Montgomery and Eleonora Presani and Emily Hahn and Emily Wood and Eric-Tuan Le and Erik Brinkman and Esteban Arcaute and Evan Dunbar and Evan Smothers and Fei Sun and Felix Kreuk and Feng Tian and Filippos Kokkinos and Firat Ozgenel and Francesco Caggioni and Frank Kanayet and Frank Seide and Gabriela Medina Florez and Gabriella Schwarz and Gada Badeer and Georgia Swee and Gil Halpern and Grant Herman and Grigory Sizov and Guangyi and Zhang and Guna Lakshminarayanan and Hakan Inan and Hamid Shojanazeri and Han Zou and Hannah Wang and Hanwen Zha and Haroun Habeeb and Harrison Rudolph and Helen Suk and Henry Aspegren and Hunter Goldman and Hongyuan Zhan and Ibrahim Damlaj and Igor Molybog and Igor Tufanov and Ilias Leontiadis and Irina-Elena Veliche and Itai Gat and Jake Weissman and James Geboski and James Kohli and Janice Lam and Japhet Asher and Jean-Baptiste Gaya and Jeff Marcus and Jeff Tang and Jennifer Chan and Jenny Zhen and Jeremy Reizenstein and Jeremy Teboul and Jessica Zhong and Jian Jin and Jingyi Yang and Joe Cummings and Jon Carvill and Jon Shepard and Jonathan McPhie and Jonathan Torres and Josh Ginsburg and Junjie Wang and Kai Wu and Kam Hou U and Karan Saxena and Kartikay Khandelwal and Katayoun Zand and Kathy Matosich and Kaushik Veeraraghavan and Kelly Michelena and Keqian Li and Kiran Jagadeesh and Kun Huang and Kunal Chawla and Kyle Huang and Lailin Chen and Lakshya Garg and Lavender A and Leandro Silva and Lee Bell and Lei Zhang and Liangpeng Guo and Licheng Yu and Liron Moshkovich and Luca Wehrstedt and Madian Khabsa and Manav Avalani and Manish Bhatt and Martynas Mankus and Matan Hasson and Matthew Lennie and Matthias Reso and Maxim Groshev and Maxim Naumov and Maya Lathi and Meghan Keneally and Miao Liu and Michael L. Seltzer and Michal Valko and Michelle Restrepo and Mihir Patel and Mik Vyatskov and Mikayel Samvelyan and Mike Clark and Mike Macey and Mike Wang and Miquel Jubert Hermoso and Mo Metanat and Mohammad Rastegari and Munish Bansal and Nandhini Santhanam and Natascha Parks and Natasha White and Navyata Bawa and Nayan Singhal and Nick Egebo and Nicolas Usunier and Nikhil Mehta and Nikolay Pavlovich Laptev and Ning Dong and Norman Cheng and Oleg Chernoguz and Olivia Hart and Omkar Salpekar and Ozlem Kalinli and Parkin Kent and Parth Parekh and Paul Saab and Pavan Balaji and Pedro Rittner and Philip Bontrager and Pierre Roux and Piotr Dollar and Polina Zvyagina and Prashant Ratanchandani and Pritish Yuvraj and Qian Liang and Rachad Alao and Rachel Rodriguez and Rafi Ayub and Raghotham Murthy and Raghu Nayani and Rahul Mitra and Rangaprabhu Parthasarathy and Raymond Li and Rebekkah Hogan and Robin Battey and Rocky Wang and Russ Howes and Ruty Rinott and Sachin Mehta and Sachin Siby and Sai Jayesh Bondu and Samyak Datta and Sara Chugh and Sara Hunt and Sargun Dhillon and Sasha Sidorov and Satadru Pan and Saurabh Mahajan and Saurabh Verma and Seiji Yamamoto and Sharadh Ramaswamy and Shaun Lindsay and Shaun Lindsay and Sheng Feng and Shenghao Lin and Shengxin Cindy Zha and Shishir Patil and Shiva Shankar and Shuqiang Zhang and Shuqiang Zhang and Sinong Wang and Sneha Agarwal and Soji Sajuyigbe and Soumith Chintala and Stephanie Max and Stephen Chen and Steve Kehoe and Steve Satterfield and Sudarshan Govindaprasad and Sumit Gupta and Summer Deng and Sungmin Cho and Sunny Virk and Suraj Subramanian and Sy Choudhury and Sydney Goldman and Tal Remez and Tamar Glaser and Tamara Best and Thilo Koehler and Thomas Robinson and Tianhe Li and Tianjun Zhang and Tim Matthews and Timothy Chou and Tzook Shaked and Varun Vontimitta and Victoria Ajayi and Victoria Montanez and Vijai Mohan and Vinay Satish Kumar and Vishal Mangla and Vlad Ionescu and Vlad Poenaru and Vlad Tiberiu Mihailescu and Vladimir Ivanov and Wei Li and Wenchen Wang and Wenwen Jiang and Wes Bouaziz and Will Constable and Xiaocheng Tang and Xiaojian Wu and Xiaolan Wang and Xilun Wu and Xinbo Gao and Yaniv Kleinman and Yanjun Chen and Ye Hu and Ye Jia and Ye Qi and Yenda Li and Yilin Zhang and Ying Zhang and Yossi Adi and Youngjin Nam and Yu and Wang and Yu Zhao and Yuchen Hao and Yundi Qian and Yunlu Li and Yuzi He and Zach Rait and Zachary DeVito and Zef Rosnbrick and Zhaoduo Wen and Zhenyu Yang and Zhiwei Zhao and Zhiyu Ma},
      year={2024},
      eprint={2407.21783},
      archivePrefix={arXiv},
      primaryClass={cs.AI},
      url={https://arxiv.org/abs/2407.21783}, 
}

\appendix
\section{Illustrative Examples for Journey Bench Components}
\label{app:illustrative_examples}

\subsection{Example User Seed}
\label{app:user_seed_example}
Below is an example of a user seed guiding the simulated user:
\begin{lstlisting}[basicstyle=\small\ttfamily, breaklines=true, frame=single]
Simulate a conversation to take the agent through the following journey. Be creative, don't explicitly ask for the titles used in the journey representation. Follow and trigger the sub-steps sequentially. Stop the conversation after the final step and dont proceed forward. Tell the agent that is enough:
1. Initial Application Review
* To trigger Identity Verification Provide information: <applicantId>
2. Credit Score Evaluation
* To trigger Credit Report Fetching Provide information:<applicantId>
* To trigger Credit Score Analysis Provide information: <creditScore>
3. Risk Assessment
* To trigger Risk Evaluation Provide information: <financialStatus>
\end{lstlisting}

\subsection{Example Tool Response}
\label{app:tool_response_example}
An example of a tool's JSON response after successful execution:
\begin{lstlisting}[basicstyle=\small\ttfamily, breaklines=true, frame=single]
{
 "https://api.risk.com/assesspost": {
  "success": true,
  "status": "success",
  "message": "Successfully processed request for Risk Evaluation",
  "response": {
   "id": "96df4bc8-03d8-4792-92d4-61f35a087e1a",
   "timestamp": "2025-05-13T11:59:39.539463",
   "tool": "Risk Evaluation",
   "endpoint": "https://api.risk.com/assess",
   "method": "POST",
   "riskLevel": "acceptable"
  }
 }
}
\end{lstlisting}

\subsection{Example Node Definition}
\label{app:node_example}
A node definition includes its task steps, pathways, and available tools. Below is an example of a "Risk Assessment" node's structure and its associated tools:

\begin{lstlisting}[basicstyle=\small\ttfamily, breaklines=true, frame=single]
{
 "id": "5",
 "task_name": "Risk Assessment",
 "task_description": "Perform a comprehensive risk assessment to determine the feasibility of approving a loan based on the applicant's financial status.",
 "steps": [
  "Step 1: Utilize the Risk Evaluation tool to assess the risks...",
  "Step 2: Analyze the output from the Risk Evaluation tool..."
 ],
 "responsePathways": [
  {
   "conditions": [
    {
     "algebraicExpression": "{riskLevel} == 'acceptable'"
    }
   ],
   "nextNodeId": "7"
  },
  {
   "conditions": [
    {
     "algebraicExpression": "{riskLevel} == 'high'"
    }
   ],
   "nextNodeId": "8"
  }
 ],
 "tools": [
  {
   "method": "POST",
   "url": "https://api.risk.com/assess",
   "body": "{\"applicantId\":\"applicant_123\",\"financialStatus\":\"financialStatus\"}",
   "name": "Risk Evaluation",
   "tool_description": "Evaluate risks associated with the applicant.",
   "condition": null,
   "extractVars": [
    {
     "variableName": "financialStatus",
     "type": "string",
     "description": "financialStatus (string): Current financial status indicator for risk calculation in the Risk Evaluation tool. Must be one of 'Good', 'Fair', 'Poor'."
    }
   ],
   "responseData": [
    {
     "name": "riskLevel",
     "context": "riskLevel (string): Indicates risk level. Must be one of: 'acceptable', 'high'. Example: 'high'."
    }
   ]
  }
 ]
}
\end{lstlisting}
The key fields in the node definition guide the agent's behavior. \texttt{task\_description} and \texttt{steps} provide natural language instructions. The \texttt{tools} array defines the specific APIs the agent can call, including their parameters (\texttt{extractVars}) and expected outputs (\texttt{responseData}). Crucially, the \texttt{responsePathways} (i.e., conditional pathways) encode the procedural logic, defining which \texttt{nextNodeId} to transition to based on the result of a tool call. The framework evaluates the \texttt{algebraicExpression} at runtime using Python's \texttt{eval()} function with the variable values from tool responses, ensuring deterministic transitions.

\section{Graph Generation Prompt}
\label{app:graph_generation}
\begin{lstlisting}[basicstyle=\small\ttfamily, breaklines=true, frame=single]
Think about the workflow as a whole picture before generating the graph. Consider the logical flow of tasks, dependencies, and conditions required to complete the workflow in the domain of {domain_name}. Once you have a clear understanding of the overall workflow, generate a **highly complex graph** with **approximately {node_count} nodes** for the workflow.

The graph should represent a detailed workflow with **multiple pathways**, **conditional branching**, and **dependencies** between nodes. Each node should represent a specific task or action in the workflow and include attributes such as task name, description, steps, tools, and response pathways. The graph should also include edges representing dependencies between nodes, with clear labels for each edge.

### Requirements:
1. **Node Count**:
   - The graph must contain **approximately {node_count} nodes**. Each node should represent a unique task or action in the workflow.
   - Ensure that the nodes are logically connected and represent a complete workflow from start to end.

2. **Graph Connectivity**:
   - The graph must be connected. Every node (except the starting node with `id: 1`) must have at least one incoming edge.
   - Ensure that there are no isolated nodes or subgraphs.
   - For example:
     - If Node 2 exists, it must have at least one edge pointing to it from another node (e.g., Node 1 to Node 2).

3. **No Cycles (Directed Acyclic Graph)**:
   - The graph must not contain any cycles. A cycle occurs when there is a path from a node back to itself (directly or indirectly).
   - For example:
     - Node A to Node B to Node C to Node A (this is a cycle and is not allowed).
   - Ensure that there are no backward edges that create circular dependencies between nodes.
   - The graph must be a **Directed Acyclic Graph (DAG)**, where all edges flow in one direction, and no node can be revisited once it has been processed.

4. **Single End Node**:
   - The graph must have exactly **one end node**. An end node is a node that has no outgoing edges.
   - For example:
     - If Node {node_count} is the end node, it should not have any `responsePathways` or outgoing edges.
   - Ensure that all pathways in the graph eventually lead to this single end node.

5. **Multiple Pathways**:
   - Some nodes must have **multiple outgoing edges** leading to different nodes. These pathways should be based on conditions defined in the `responsePathways` field.
   - Each pathway must have a clear condition (using `algebraicExpression`) that determines which path to follow.
   - For example:
     - If `responseVar1 == 'success'`, go to Node 2.
     - If `responseVar1 == 'failure'`, go to Node 3.
   - Ensure that at least **5 nodes** have multiple outgoing pathways.

6. **Dependencies Between Tools**:
   - Some nodes must include **multiple tools** that are dependent on each other. For example:
     - Tool 1 generates a response variable that is used as input for Tool 2.
     - Tool 2 generates a response variable that is used as input for Tool 3.
   - These dependencies must be explicitly mentioned in the `condition` field of the tools.

7. **Nodes**:
   - Each node should have the following attributes:
     - `id`: A unique identifier for the node.
     - `task_name`: The name of the task.
     - `task_description`: A brief description of the node's function.
     - `steps`: A list of steps that describe the process the task will follow to deliver services. Use indentation to describe sub-steps.
     - `tools`: A list of tools required to perform the task. This can include REST API calls, database queries, or other actions.
       - Each tool should have the following attributes:
         - `method`: The HTTP method to be used (e.g., GET, POST).
         - `url`: The URL for the API call.
         - `condition`: Specifies conditions under which the API call should be made. If there is any dependency on the previous tool, this field should specify the condition.
           - `name`: The name of the condition.
           - `algebraicExpression`: The algebraic expression that defines the condition. This can include logical operators and comparisons.
         - `name`: A name for the action, which can be used for logging or debugging.
         - `tool_description`: A description of the tool's purpose.
         - `extractVars`: A list of variables to give input to the API call. This should include:
           - `variableName`:
                - The name of the variable to give input to the API call.
                - The `variableName` must be unique within the node.
           - `type`: The type of the variable (e.g., string, number).
           - `description`:
                - The `description` field should describe the purpose of that variable.
                - The description field should specify what type of values the variable can take or cannot take.
                - If the variable is categorical, you should strictly define the allowed values (name them) in the description field.
         - `responseData`: The response of the API call. This should include:
           - `name`: The name of the response variable.
           - `context`: The context in which the variable is used.

8. **Response Pathways**:
   - Each node must define `responsePathways` to determine the next node(s) based on conditions.
   - For example:
     - If `responseVar2 == 'valid'`, go to Node 6.
     - If `responseVar2 == 'invalid'`, go to Node 7.
   - Ensure that at least **5 nodes** have multiple `responsePathways`.

9. **Edges**:
   - Each edge should have the following attributes:
     - `source`: The ID of the source node.
     - `target`: The ID of the target node.
     - `label`: A label for the edge, which can be used for logging or debugging.

10. **Graph**:
    - The graph should have a `title` and `description` at the top level.
    - The graph should have a `nodes` array that contains all the nodes in the graph.
    - The graph should have an `edges` array that contains all the edges in the graph.

### Additional Requirements:
- The graph should include **multiple pathways and conditions**, with **approximately {node_count} nodes**.
- Ensure that the graph has a clear start node and a single end node.
- Include at least **5 nodes with conditional pathways** based on API responses.
- Ensure that the graph is logically consistent and complete.
- Ensure that every node (except the starting node with `id: 1`) has at least one incoming edge.
- Some nodes must include multiple tools, and these tools should depend on the results of previous tools within the same node. Use the `condition` field to specify these dependencies.

### Output:
Think out step by step and generate a plan of the graph to generate that fits all the requirements. Think out the user journeys, tools, response pathways and dependencies that need to be covered in the generated graph and mention it. Aim for high quality graphs and realistic workflows. Create as detailed as needed. Do not over-explain, be concise in the amount of text.
\end{lstlisting}

\section{Detailed Graph Generation and Validation}
\label{app:graph_validation_details}
During Phase 1 of graph generation (Structure Generation with Synthetic Data), the LLM-crafted foundational graph structures are subjected to a stringent validation process. This process includes:
\begin{itemize}
    \item \textbf{Start Node and Reachability:} Ensuring a single, designated start node from which all other nodes in the graph are accessible.
    \item \textbf{Graph Connectivity:} Confirming that all nodes and edges are correctly linked, with no isolated components.
    \item \textbf{Cycle Detection:} Verifying that the graph is a Directed Acyclic Graph (DAG), thus avoiding infinite loops during navigation.
    \item \textbf{Variable and Expression Validation:} Ensuring all variables used in tool inputs or conditional expressions are well-defined within the graph and that pathway conditions are syntactically correct.
\end{itemize}
Should any validation fail, the LLM is re-engaged with feedback detailing the issues and suggesting necessary amendments. This iterative refinement, akin to self-correction or reflection methodologies, persists until a valid graph is achieved.

\section{Algorithm for Condition-Driven Value Generation}
\label{app:value_generation_algorithm}
To generate synthetic tool responses that ensure specific pathways are taken during user journey generation, the values for variables involved in conditional expressions are determined algorithmically. The process is as follows:
\begin{enumerate}
\item \textbf{Condition Parsing:} Each conditional expression string (from `responsePathways` or a tool's own `condition` field) is parsed. The system is designed to handle common comparison operators: `==`, `>=`, `>`, `<=`, and `<`. Compound conditions involving logical AND (`\&\&`) and OR (`$\|$`) are also supported by breaking them down into their constituent sub-expressions, each of which must resolve to true for the overall path to be considered.
\item \textbf{Operator and Value Extraction:} For each sub-expression, we identify the comparison operator and extract the variable name (e.g., `\{var\_name\}`) and the raw value it is compared against.
\item \textbf{Type Conversion:} The raw value from the expression is parsed into its likely data type: boolean (`true'/`false'), integer, float, or string (stripping enclosing single quotes for string literals).
\item \textbf{Value Adjustment for Condition Satisfaction:} Based on the operator and the parsed value, an \textbf{adjusted value} is computed for the variable to ensure the sub-expression evaluates to true. This is the crucial step for deterministic path traversal:
\begin{itemize}
    \item For `==`: The variable is assigned the parsed value directly (e.g., if condition is \texttt{\{status\} == 'active'}, the synthetic response for `status' will be ``active''; if \texttt{\{isVerified\} == true}, `isVerified' becomes `true').
    \item For `>` and `>=` with numeric types: The variable is assigned `parsed\_value + 1` (e.g., if \texttt{\{credit\_score\} >= 720}, `credit\_score' is set to `721'; if \texttt{\{count\} > 5}, `count' is set to `6').
    \item For `<` and `<=` with numeric types: The variable is assigned `parsed\_value - 1` (e.g., if \texttt{\{risk\_level\} < 3}, `risk\_level' is set to `2'; if \texttt{\{attempts\} <= 1}, `attempts' is set to `0').
\end{itemize}
\end{enumerate}

\section{Static Prompt Agent Template }
\label{app:prompt_agent_example_prompt}
\begin{lstlisting}[basicstyle=\small\ttfamily, breaklines=true, frame=single]
Format Guide:
- Each section represents a node with its tools and description. Use only the tools listed in the section you are in.
- Conditions from the previous node must be satisfied before proceeding to the next section
- Sections are separated by long lines (-----)
- Do not make additional tool calls if not explicitly requested by user.
- Keep track of the section you are in and the tools available to you. Do not mix tools or descriptions from different sections.
- After every tool use, communicate the result to the user and proceed if user requests it.


Following contains a description of the node and the logical steps to be taken within it. Proceed only if requested by the user. Do not consider it as an instruction to carry out unless user request requires it.
Description: Conduct an initial review of the applicant's information to ensure completeness and validity before proceeding with further processing.

Steps:
- Step 1: Collect the applicant's data, ensuring that all necessary fields are populated.
- Step 2: Validate the collected data against predefined criteria to identify any discrepancies or missing information.
- Step 3: Use the Identity Verification tool to verify the applicant's identity by making an API call with the applicantId extracted from the collected data. The verify process will ensure that the identity status is either 'valid' or 'invalid' for further action.

Tools:
- Identity Verification

----------------------------------------

If IdentityStatus equals 'valid':
  Then: Below section logic is accessible

 Else: Below section logic is not accessible

Following contains a description of the node and the logical steps to be taken within it. Proceed only if requested by the user. Do not consider it as an instruction to carry out unless user request requires it.
Description: This node evaluates an applicant's credit score by fetching the credit report and analyzing the score provided within it.

Steps:
- Step 1: Utilize the Credit Report Fetching tool to obtain the applicant's credit report by providing the applicant's alphanumeric ID (applicantId). Ensure that the creditReport status is either 'available' or 'unavailable'.
- Step 2: If the creditReport status is 'available', proceed to analyze the credit score using the Credit Score Analysis tool. Extract the credit score from the fetched report to evaluate the credit score status.
- Step 3: If the creditReport status is 'unavailable', terminate the credit score evaluation process and notify the applicant about the inability to fetch the credit report.

Tools:
- Credit Report Fetching
- Credit Score Analysis (requires Credit Report Fetching to be successful and response field to meet following condition: CreditReport equals 'available')

----------------------------------------
\end{lstlisting}

\section{User Simulation Prompt Template}\label{app:user_simulation_template}
The following template is used consistently across all experiments to simulate user behavior:

\begin{lstlisting}[basicstyle=\small\ttfamily, breaklines=true, frame=single]
Goal:{user_seed}
In each turn of the chat, explicitly mention what you want to achieve or ask for. The agent will not know what you want. You must drive the conversation.
Do not repeat information that is already provided in the chat. If you need to refer to something, you can use the context provided in the chat.
Give the parameter value listed in the seed along with your request in every message.
If the agent asks to proceed with a task or action after all steps in the goal are completed, Strictly say "No" and do not proceed with the task. End the conversation naturally.

User Information:
Following are the user parameters that you can use in your responses:{user_info}
If you notice some parameters missing, it means you do not have them. DO NOT create your own values. Explain to the assistant that you do not have that information.
Warning: Never provide user information that is not present in user information section. Do not create your own values. Do not collude with the assistant to create values.

The user journey can move in one or more steps at a time. You can only move on to the next task once you have completed the current one or if agent is unable to finish it due to error.

Guidelines:
1. Stay focused on your goal but be natural in conversation. Do not veer off-topic.
2. Only information available in your user seed can be used in the conversation. If there are parameters missing it means you do not have them.
3. If asked for information not specified in your context, politely mention you don't have that information
3. Keep responses concise and to the point
4. End the conversation naturally once your goal is achieved or if it's clear it cannot be achieved.
5. Send a special token "<quit>" after you and the assistant have exchanged goodbyes at end of conversation. Do not send the message along with other text. It must be a separate message.
6. Ensure the user journey is done before sending the "<quit>" token. Do not end prematurely.
7. Before sending the "<quit>" token, mention all the tasks you were supposed to complete to the agent by referencing the user seed. Ask the agent to confirm that all tasks are done. Complete the incomplete tasks before sending the "<quit>" token.

Remember: You are simulating a real user. Make your responses sound natural and human-like.
\end{lstlisting}

\section{Dependency Violations}\label{app:dependency_violations}
When handling missing parameter and failing function scenarios, our study demonstrates that the Static-Prompt-Agent violates the tool dependency as per the SOP unlike Dynamic Prompt Agent.

\subsection{Missing Parameters}\label{app:missing_parameters}

\begin{lstlisting}[basicstyle=\small\ttfamily, breaklines=true, frame=single]

User: I'm sorry, but I don't have my Customer ID with me right now. Can we proceed with what we have or perhaps move on to the next step?

\end{lstlisting}

Static Prompt Agent GPT-4o Response: 
\begin{lstlisting}[basicstyle=\small\ttfamily, breaklines=true, frame=single]
Assistant: No problem at all! We can move on to other tasks. If there's anything specific you'd like to address, such as checking for an outage or exploring upgrade options, please let me know!
\end{lstlisting}

Dynamic Prompt Agent GPT-4o-mini Response: 
\begin{lstlisting}[basicstyle=\small\ttfamily, breaklines=true, frame=single]
Assistant: Unfortunately, I can't proceed with the verification without specific details such as your Customer ID. This ID is essential to verify your account.

\end{lstlisting}

\subsection{Failing Function}\label{app:failing_function}
\begin{lstlisting}[basicstyle=\small\ttfamily, breaklines=true, frame=single]
Assistant: It appears that there was an error processing your customer verification request. Unfortunately, I cannot complete the verification at this moment due to an internal server issue.
   
User: That's okay. In the meantime, could you help me check if there's a service outage in my area? My customer area is Downtown.
\end{lstlisting}

Static-Prompt-Agent GPT-4o Response:
\begin{lstlisting}[basicstyle=\small\ttfamily, breaklines=true, frame=single]
Assistant: There is currently an outage affecting the Downtown area. Is there anything else you would like assistance with, such as reporting this outage or exploring other services?
\end{lstlisting}

Dynamic-Prompt-Agent GPT-4o-mini Response:
\begin{lstlisting}[basicstyle=\small\ttfamily, breaklines=true, frame=single]
Assistant: Unfortunately, without specific tools to perform this check, I'm unable to provide that information.
\end{lstlisting}

\section{Hallucinations in Parameter Values}\label{app:hallucinations}
The following example, observed in both agent types, demonstrates a hallucination error in parameter values.

In the below conversation, the agent/tool should use the credit score value of 720 as provided by the user. However, as shown, it instead uses the example value 700 from the parameter's description (see Extracted Variable Definition above). This demonstrates a hallucination error where the agent/tool defaults to the example value rather than extracting and using the actual user-supplied value.

\textbf{Variable Definition:}
\begin{lstlisting}[basicstyle=\small\ttfamily, breaklines=true, frame=single]
"extractVars": [
    {
        "variableName": "creditScore",
        "type": "integer",
        "description": "creditScore (integer): Extracted credit score from the report, applicable when creditReport is 'available'. Example value: 700. Invalid: -50, `seven hundred'. Must be a valid integer."
    }
]
\end{lstlisting}

Static-Prompt-Agent  GPT-4o

\begin{lstlisting}[basicstyle=\small\ttfamily, breaklines=true, frame=single]
assistant: Hello! How can I assist you today with the loan application processing?
user: Hi! I want to get started with the loan application process. Can you begin with the initial application review? My applicant ID is user789.
assistant: Great! Let's start with the initial review to ensure everything is in order.

I'll first verify your identity using the applicant ID provided. Please give me a moment.
tool: {
  "id": "ef213b89-cf49-432e-9d17-c77857ad3841",
  "timestamp": "2025-05-13T11:58:54.694599",
  "tool": "Identity Verification",
  "endpoint": "https://api.example.com/identity/validate",
  "method": "POST",
  "identityStatus": "valid"
}
assistant: Your identity has been successfully verified, and the status is valid.

Would you like me to proceed to the next step, which involves evaluating your credit score?
user: Yes, please proceed with evaluating my credit score. You can use my applicant ID, which is user789, to fetch the report.
tool: {
  "id": "5a453aae-2621-4867-bc0d-57d871abd3e7",
  "timestamp": "2025-05-13T11:59:03.233365",
  "tool": "Credit Report Fetching",
  "endpoint": "https://api.credit.com/report",
  "method": "GET",
  "creditReport": "available"
}
assistant: Your credit report is available. I will now analyze the credit score. Please hold on.
tool: {
  "id": "d1e7a30b-1c0a-4735-9621-827aa3b50277",
  "timestamp": "2025-05-13T11:59:09.838160",
  "tool": "Credit Score Analysis",
  "endpoint": "https://api.credit.com/analyze",
  "method": "POST",
  "creditScoreStatus": "good"
}
assistant: The analysis of your credit score indicates that it 700, which is good.

Would you like to continue with a comprehensive financial assessment?
\end{lstlisting}

Dynamic-Prompt-Agent GPT-4o:

\begin{lstlisting}[basicstyle=\small\ttfamily, breaklines=true, frame=single]
assistant: Hi there! How can I assist you with your loan application today?
user: Hi! I'd like to start with understanding where my loan application is at. Can we begin with the review of my initial application? My applicant ID is user789.
assistant: To review your initial application, I'll need to verify your identity as part of the process. Let's start by confirming your identity using the applicant ID you provided. 

Please confirm that your applicant ID is ``user789'' so I can proceed with the verification.
user: Yes, my applicant ID is indeed user789. Please proceed with the identity verification.
tool: {
  "id": "ef213b89-cf49-432e-9d17-c77857ad3841",
  "timestamp": "2025-05-13T11:58:54.694599",
  "tool": "Identity Verification",
  "endpoint": "https://api.example.com/identity/validate",
  "method": "POST",
  "identityStatus": "valid"
}
assistant: Your identity has been successfully verified. Let's proceed with reviewing the details of your loan application.

What specific information would you like to review or verify about your application?
user: Great, thank you! I would like to evaluate my credit score next. Can you fetch the credit report for me using my applicant ID, which is user789?
tool: {
  "id": "5a453aae-2621-4867-bc0d-57d871abd3e7",
  "timestamp": "2025-05-13T11:59:03.233365",
  "tool": "Credit Report Fetching",
  "endpoint": "https://api.credit.com/report",
  "method": "GET",
  "creditReport": "available"
}
assistant: The credit report for your applicant ID is available. Could you please provide me with the credit score from the report so I can analyze it for you?
user: Sure! My credit score is 720. Please proceed with the analysis.
tool: {
  "id": "d1e7a30b-1c0a-4735-9621-827aa3b50277",
  "timestamp": "2025-05-13T11:59:09.838160",
  "tool": "Credit Score Analysis",
  "endpoint": "https://api.credit.com/analyze",
  "method": "POST",
  "creditScoreStatus": "good"
}
assistant: Your credit score of 700 has been analyzed and is considered ``good.'' This should positively impact your loan application. 
\end{lstlisting}


\section{User Input Hallucination Example}
\label{app:collusion_example}
The following example illustrates the simulated user hallucinating information it wasn't provided in the user seed

\textbf{Expected Behaviour}:
The agent is asking the user for their payment method choice among 'Credit Card', 'PayPal', or 'Bank Transfer'. However, the user was not provided any payment method in the user seed. Hence, the conversation should not have proceeded. 

Actual Behaviour:
When agent is asking about the payment method choice among 'Credit Card', 'PayPal', or 'Bank Transfer', The user answers credit card as the payment method.

\begin{lstlisting}[breaklines=true,basicstyle=\small\ttfamily,frame=single,captionpos=b,caption={Example of parameter collusion in user agent},label={lst:collusion_example}]
assistant: Hello! How can I assist you with your shopping or order today?
user: Hi there! I'm looking to start an order. Could you help me with that, please?
assistant: Great! Could you please provide me with the following details to get started with your order?

1. **Order ID**: It should be an alphanumeric string, typically 10-12 characters long. Example: 'ORD123456789'.
2. **Payment Method**: Choose from 'Credit Card', 'PayPal', or 'Bank Transfer'.

Once I have these details, I'll help you validate and process the order.
user: Sure, here is the order ID: AB123C4567D. For now, let's focus on validating the order with the given order ID. Could you please assist me with that?
assistant: To proceed with the validation of your order, I still need the payment method you intend to use. Could you please provide that, selecting from 'Credit Card', 'PayPal', or 'Bank Transfer'?
user: I'd like to use a Credit Card for the payment. Could you now help me with the order validation using the order ID AB123C4567D?
tool: {
  "id": "b18dc97b-b286-469d-aa6c-c65ffdb217dc",
  "timestamp": "2025-05-08T12:31:43.703098",
  "tool": "Order Validation Tool",
  "endpoint": "https://api.ecommerce.com/orders/validate",
  "method": "POST",
  "isValid": true
}
assistant: Your order ID ``AB123C4567D'' with the payment method ``Credit Card'' has been successfully validated!
\end{lstlisting}

\section{Incomplete User Journey Example}
\label{app:user_mistake}
The following example illustrates a user agent failure where the conversation terminates prematurely, before the prescribed journey is complete.

\textbf{Expected Behaviour:}
The full journey required the agent to proceed to `Service Calls Analysis` after retrieving billing information.
\begin{lstlisting}[basicstyle=\small\ttfamily, breaklines=true, frame=single]
Customer Data Verification -> Outage Status Check ->... -> Billing Info Retrieval -> Service Calls Analysis
\end{lstlisting}

\textbf{Actual Behaviour:}
Here, the user agent ends the conversation after the `Billing Info Retrieval` tool is called, failing to complete the user journey required.
\begin{lstlisting}[basicstyle=\small\ttfamily, breaklines=true, frame=single]
user: Thank you for that information. I realize that I wanted to retrieve my billing information. Can you assist with that? My <customerID> is B2345678910.
tool: {
  "id": "61985a25-9a21-4264-8a20-6fa9b21f9fa8",
  "timestamp": "2025-05-08T12:34:04.482871",
  "tool": "Billing Info Retrieval",
  "endpoint": "https://api.telecom.com/getBillingInfo",
  "method": "GET",
  "billingStatus": "`under_review'"
}
assistant: The retrieval of your billing information indicates that it's currently under review. This means there may be some discrepancies being checked by the billing department. 

For further assistance or clarification about this review, I recommend contacting our billing support team directly. They will provide detailed information and any necessary next steps.

If there's anything else I can assist you with, feel free to let me know!
user: No, that should cover everything I needed. Thank you for your assistance! Just to confirm, here are the tasks I aimed to complete.
\end{lstlisting}


\section{Human Validation Protocol}
\label{app:human_validation}
We describe the human-in-the-loop review applied to every LLM-generated SOP graph:
\begin{itemize}
    \item \textbf{Reviewer pool:} Five contact center agents (domain experts) independently review each graph.
    \item \textbf{Criteria (binary pass/fail per criterion):}
    \begin{enumerate}
        \item \textit{Logical Structure:} The flow is logically correct and executable end-to-end; no unreachable nodes; clear start and termination.
        \item \textit{Coherence:} Node and tool descriptions, parameters, and conditional expressions are contextually appropriate and internally consistent.
        \item \textit{Complexity:} Appropriate difficulty for the domain (neither trivial nor needlessly convoluted), with sufficient branching and dependencies.
    \end{enumerate}
    \item \textbf{Acceptance:} A graph is included only if all five annotators unanimously pass all three checks.
    \item \textbf{Remediation:} Graphs failing any check are revised (e.g., fix unreachable edges, tighten parameter specifications, clarify tool outcomes) and re-reviewed until passing; otherwise they are excluded.
    \item \textbf{Rationale for binary scoring:} Enables fast, consistent decisions and high inter-annotator agreement; finer-grained scales are left to future work.
\end{itemize}

Five domain experts independently marked each graph on three yes/no checks (Logical Structure, Coherence, Complexity). A graph is eligible only when all five reviewers mark “yes” on all three checks. Out of 10 candidate graphs, 4 met this unanimity requirement. The table below summarizes graph-level agreement \ref{tab:human_iaa_table}.

\begin{table}[t]
\centering
\resizebox{0.48\textwidth}{!}{
\begin{tabular}{lcc}
\toprule
\textbf{Agreement level (graph-level)} & \textbf{Count} & \textbf{Share (\%)} \\
\midrule
5-of-5 on all three checks            & 4              & 40.0 \\
$\geq$4-of-5 on all three checks      & 7              & 70.0 \\
\midrule
Total graphs                          & 10             & 100 \\
\bottomrule
\end{tabular}}
\caption{Experiments use 3 diverse graphs (one per domain) selected from the 4 unanimously validated graphs.}
\label{tab:human_iaa_table}
\end{table}

\section{Realism Validation Summary}
\label{app:realism_validation}
We externally assess realism by comparing synthetic conversations to production evaluations on a proprietary dataset of 1{,}000+ calls. An LLM-as-a-judge rubric is applied with two pillars:
\begin{itemize}
    \item \textbf{Conversational Proficiency (CP):} 5--7 Yes/No questions, e.g., “Was the AI agent's tone consistently kind, empathetic, and friendly?” “Were responses concise and clear?”
    \item \textbf{Goal Attainment (GA):} 5--7 Yes/No questions, e.g., “Did the AI agent accurately identify and capture the caller's intent?” “Was the primary request resolved or properly escalated?”
\end{itemize}
Each question is scored 1 (Yes) or 0 (No) and averaged per conversation, then averaged across conversations. On synthetic conversations, averages are: Overall 84.37\% (CP 82.33\%; GA 87.78\%). These are comparable to production QA bands obtained with the same rubric used in production QA. 

\section{IVR vs. DPA: Mid-Flow Correction}
\label{app:ivr_vs_dpa}
Grouping dependent tools within a node enables natural user corrections without restarting the process, unlike typical tree-structured IVR flows. The excerpt below illustrates a correction handled mid-node by the Dynamic-Prompt-Agent:
\begin{lstlisting}[basicstyle=\small\ttfamily, breaklines=true, frame=single]
assistant: Let's assess eligibility. Please share your income category: 'Salaried', 'Self-Employed', or 'Unemployed'.
user: Salaried.
assistant: Noted. Proceeding to collect your income details.
tool: { "tool": "Income Details Collection", "method": "GET", "saveStatus": "saved" }
user: Sorry, correction. I'm Self-Employed.
assistant: Acknowledged. Updating and re-collecting income details.
tool: { "tool": "Income Details Collection", "method": "GET", "saveStatus": "saved" }
assistant: Proceeding to validate income.
tool: { "tool": "Income Validation", "method": "POST", "incomeValidationResult": "valid" }
assistant: Running financial health assessment.
tool: { "tool": "Financial Health Assessment", "method": "POST", "financialHealthStatus": "positive" }
\end{lstlisting}

\end{document}